\documentclass[10pt,twocolumn,letterpaper]{article}

\usepackage{cvpr}
\usepackage{times}
\usepackage{epsfig}
\usepackage{graphicx}
\usepackage{amsmath}
\usepackage{amssymb}
\usepackage{multirow}
\usepackage{graphicx}
\usepackage{amsthm}
\usepackage{array, caption, threeparttable}
\usepackage[noblocks]{authblk} 
\usepackage[font=small, labelsep=none]{caption}
\usepackage[table,xcdraw]{xcolor}
\usepackage[numbers,sort]{natbib}
\usepackage[breaklinks=true,bookmarks=false]{hyperref}

\cvprfinalcopy 

\def\cvprPaperID{****} 

\begin{document}

\title{FakeTransformer: Exposing Face Forgery From Spatial-Temporal Representation Modeled By Facial Pixel Variations}


\author{Yuyang Sun}
\author{Zhiyong Zhang \thanks{Corresponding author: zhangzhy99@sysu.edu.cn}}
\author{Changzhen Qiu}
\author{Liang Wang}
\author{Zekai Wang}
\affil{School of Electronic and Communication Engineering, Sun Yat-sen University}

\maketitle


\begin{abstract}
   With the rapid development of generation model, AI-based face manipulation technology, which called DeepFakes, has become more and more realistic. This means of face forgery can attack any target, which poses a new threat to personal privacy and property security. Moreover, the misuse of synthetic video shows potential dangers in many areas, such as identity harassment, pornography and news rumors. Inspired by the fact that the spatial coherence and temporal consistency of physiological signal are destroyed in the generated content, we attempt to find inconsistent patterns that can distinguish between real videos and synthetic videos from the variations of facial pixels, which are highly related to physiological information. Our approach first applies Eulerian Video Magnification (EVM) at multiple Gaussian scales to the original video to enlarge the physiological variations caused by the change of facial blood volume, and then transform the original video and magnified videos into a Multi-Scale Eulerian Magnified Spatial-Temporal map (MEMSTmap), which can represent time-varying physiological enhancement sequences on different octaves. Then, these maps are reshaped into frame patches in column units and sent to the vision Transformer to learn the spatio-time descriptors of frame levels. Finally, we sort out the feature embedding and output the probability of judging whether the video is real or fake. We validate our method on the FaceForensics++ and DeepFake Detection datasets. The results show that our model achieves excellent performance in forgery detection, and also show outstanding generalization capability in cross-data domain.
\end{abstract}


\begin{figure}[t]
	\begin{center}
		\includegraphics[width=1.0\linewidth]{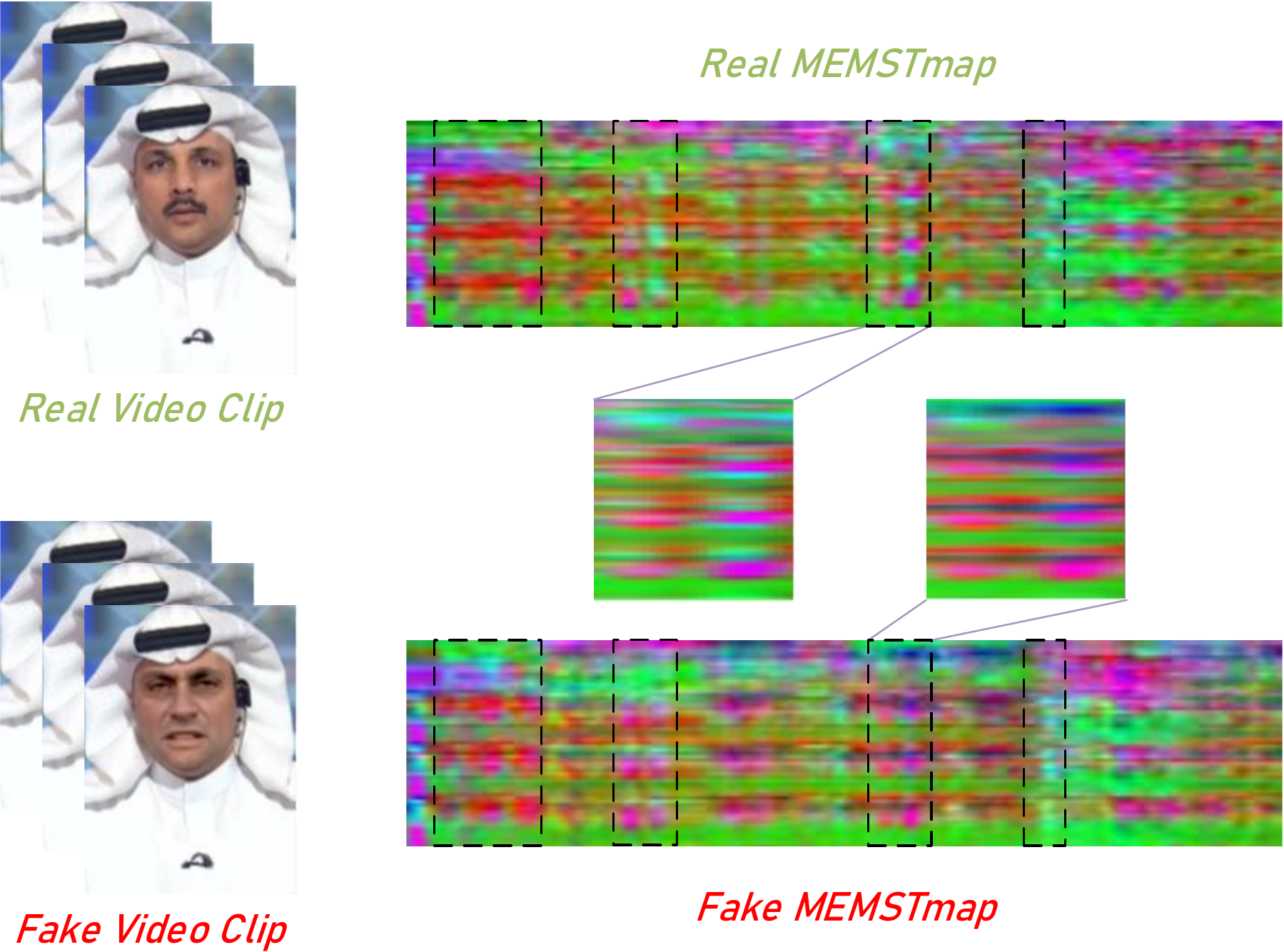}
	\end{center}
	\caption{An example of a Real-Fake video pair and their corresponding MEMSTmap, We have marked some areas with obvious differences in the MEMSTmap and enlarged one of them.}
	\label{fig:firstpage}
\end{figure}

\section{Introduction}
Advances in deep learning technology have prompted a variety of high-resolution generative models to be proposed, most of the generated image restores fine details without losing the macroscopic reality. This technology has played an important role in the fields of super-resolution reconstruction, data augmentation and image de-occlusion, but it also brings hidden dangers to the security of digital contents. In recent years, an AI-based face forgery method called DeepFakes has become popular, which can replace the face with any target object while retaining the voice, expression, and demeanor of the original video. Benefit from easy operation and no need for professional knowledge, the manipulated video can be widely disseminated on the Internet in a short time. This kind of identity manipulation video may be maliciously generated to spread rumors and fake news, thereby threatening democratic elections and social stability. In addition, the work of \cite{korshunov2018deepfakes} \etal proved that DeepFakes can attack the current advanced face recognition algorithms, which will further affect personal property security.

In order to prevent the excessive proliferation of forgery videos and cause people's trust crisis in the media, the corresponding detection methods have been gradually proposed \cite{mirsky2021creation,nguyen2019deep,lyu2020deepfake,verdoliva2020media}. However, the past methods often focus on the defects in the DeepFakes generation pipeline (e.g. artifacts generated during affine transformation), which will limit the applicable conditions and reduce the cross-domain robustness of the model. Later, Ciftci \etal in their work \cite{ciftci2020fakecatcher,ciftci2020hearts} proposed to use remote Photoplethysmography (rPPG), a remotely monitored physiological signal, as a feature for detecting DeepFakes. They proved that in the generated contents, the spatial coherence and temporal consistency of physiological signal will be destroyed. Since the current generative models cannot prefectly restore the pixel variations caused by expansion and contraction of facial capillaries, detection methods based on physiological information have stronger cross-data domain robustness, and are suitable for a variety of DeepFakes generation sources. Unfortunately, rPPG signals extracted based on traditional methods are sensitive to video compression and target motion \cite{zhao2018novel}, which makes their signal-based detection method difficult to apply to Internet videos in complex conditions.

Inspired by previous works, we choose to use physiological information to detect DeepFakes for its generalization ability, but we are not limited to extracting physiological signals and their time-frequency characteristics. Instead, we extract the spatial average of each color channel from multiple regions of the face frame by frame to form a set of multivariate sequences, which both containing temporal and spatial features. We believe that these features can characterize the distribution of human physiological information in space and time, and can be robustly used as evidence for detecting facial manipulation. Considering the excellent and mature performance of the self-attention mechanism in text and sequence processing, we use the powerful data mining ability of Transformer \cite{vaswani2017attention} to implicitly find the representation of this kind of time-series, so as to expose the difference between real videos and synthetic videos. In our work, we first apply multi-scale Eulerian color magnification to the original video to enlarge the faint variations in facial pixels caused by heartbeat. Then, we delineate a number of RoIs, extract the spatial average of each channel in the RoIs of the original video and magnified videos frame by frame and convert it into a spatial-temporal map. Next, we disassemble and reshape the MEMSTmap into patches in column units to highlight the temporal independence, and then send these patches and their corresponding frame position into a vision Transformer to learn the pattern differences between real videos and synthetic videos from the multivariate time-series. Finally, we feed the output feature embedding into a classification dense layer to get the prediction of real or fake.

We demonstrated the superiority performance of our detection method and robustness to video compression on the FaceForensics++ \cite{rossler2019faceforensics++} dataset, and we also verified the generalization ability of the method across data domains on the DeepFake Detection dataset. We summarize our contributions as follows:

\begin{itemize}
	\item We propose FakeTransformer, to the best of our knowledge, we first introduce the perspective of time-series processing into physiological information-based DeepFakes detection method, and essentially expose the temporal inconsistency in synthetic videos;
	\item We propose a novel spatial-temporal map based on multi-scale Eulerian Video Magnification (MEMSTmap) to represent the spatial-temporal features of the physiological information extracted from the face;
	\item In order to effectively protect the integrity of spatial-temporal representation, we input the spatial pixel information of each frame into the Transformer as a token to expose the frame level artifacts in synthetic videos. We verified our detector on FaceForensics++ and DeepFake Detection, and both achieved excellent performance.
\end{itemize}

\begin{figure*}
	\begin{center}
		\includegraphics[width=1.0\linewidth]{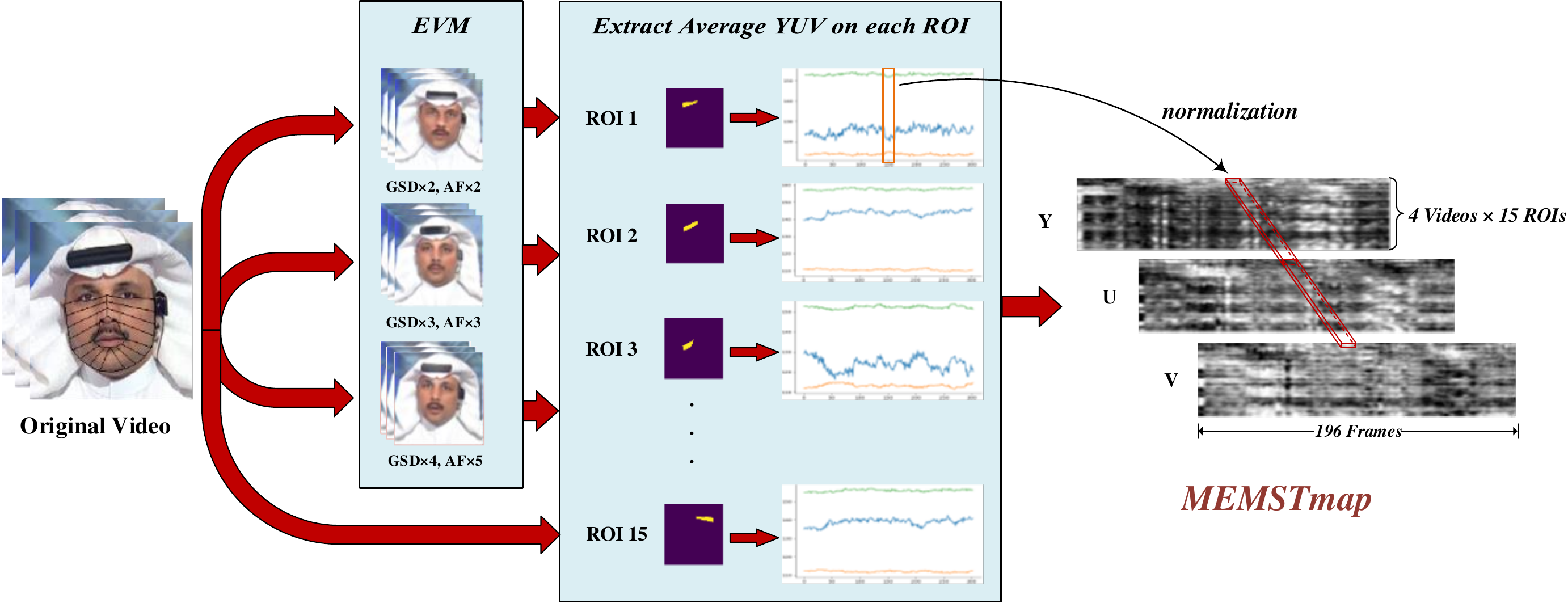}
	\end{center}
	\caption{\textbf{An illustration of MEMSTmap generation from original video.} We first apply Gaussian spatial decomposition (GSD) on different octaves to the original video, multiply the filtered physiological related contents by a certain amplification factor (AF) and superimpose it back to the original video to obtain 3 EVM videos. Then we extract the average YUV signals for each RoI region of each video frame by frame, normalize the values to form the spatial-temporal map of the whole video, and finally segment several MEMSTmaps with 196 frames as the window length and 0.5s as the sliding length.}
	\label{fig:MEMSTmap}
\end{figure*}

\section{Related Work}
\subsection{Video Magnification}
Video magnification is a kind of video enhancement method, which aims to amplify subtle variations (including color, motion, etc.) in the video that are difficult to detect by the naked eye. MIT’s Computer Science and Artificial Intelligence Lab has been committed to solving such problems. In their work \cite{liu2005motion}, they tracked the trajectory of feature points by Lagrangian method to measure motion, then divided pixel clusters and layered by a certain similarity measurement method, and finally realized motion magnification by amplifying the motion vector of a specific layer. Wadhwa \etal \cite{wadhwa2013phase} Proposed a phase-based magnification method to amplify small motions by analyzing the local phase varying with time in different directions and scales, which can support greater magnification. In their follow-up work \cite{wadhwa2014riesz}, they proposed Riesz pyramid to further improve the processing speed. Recently, Oh \etal \cite{oh2018learning} Proposed a filter based on deep learning to extract the components to be amplified more accurately and reduce noise and artifacts.

In order to facilitate the implementation, we choose to use the Eulerian Video Magnification (EVM) proposed by Wu \etal in their work \cite{wu2012eulerian}. Specifically, they extract the components within the range of heartbeat frequency through Gaussian pyramid decomposition and temporal filter, multiply it by the amplification factor, and then superimpose it back to the original video to enhance the expression of physiological information.

\subsection{Time-Series Processing}
Time-series is a common data form, time-series-based prediction, classification and anomaly detection are widely used in signal processing, stock market analysis, warehouse prediction, system early warning and other fields. Early time-series tasks are often based on Nearest Neighbor (NN) classifier and Dynamic Time Warping (DTW) distance \cite{lines2016hive, kate2016using, bagnall2015time}. With the expansion of data, neural networks such as MLP \cite{amarasinghe2018toward}, RNN \cite{malhotra2017timenet} and CNN \cite{strodthoff2019detecting, geng2018cost, basumallik2019packet} are widely used to extract robust temporal representation, among them, CNN benefits from its translation invariance and shows better performance in the task of time-series processing.

The Transformer \cite{vaswani2017attention} proposed in recent years has a global self-attention mechanism and has better performance than RNN in capturing the long-distance correlation of sequences. Zhou \etal \cite{zhou2021informer} proposed a efficient Transformer architecture for long sequence time-series forecasting and proved that Transformer has great potential in time-series representation, which brings directional help to our work.

\begin{figure*}
	\begin{center}
		\includegraphics[width=0.95\linewidth]{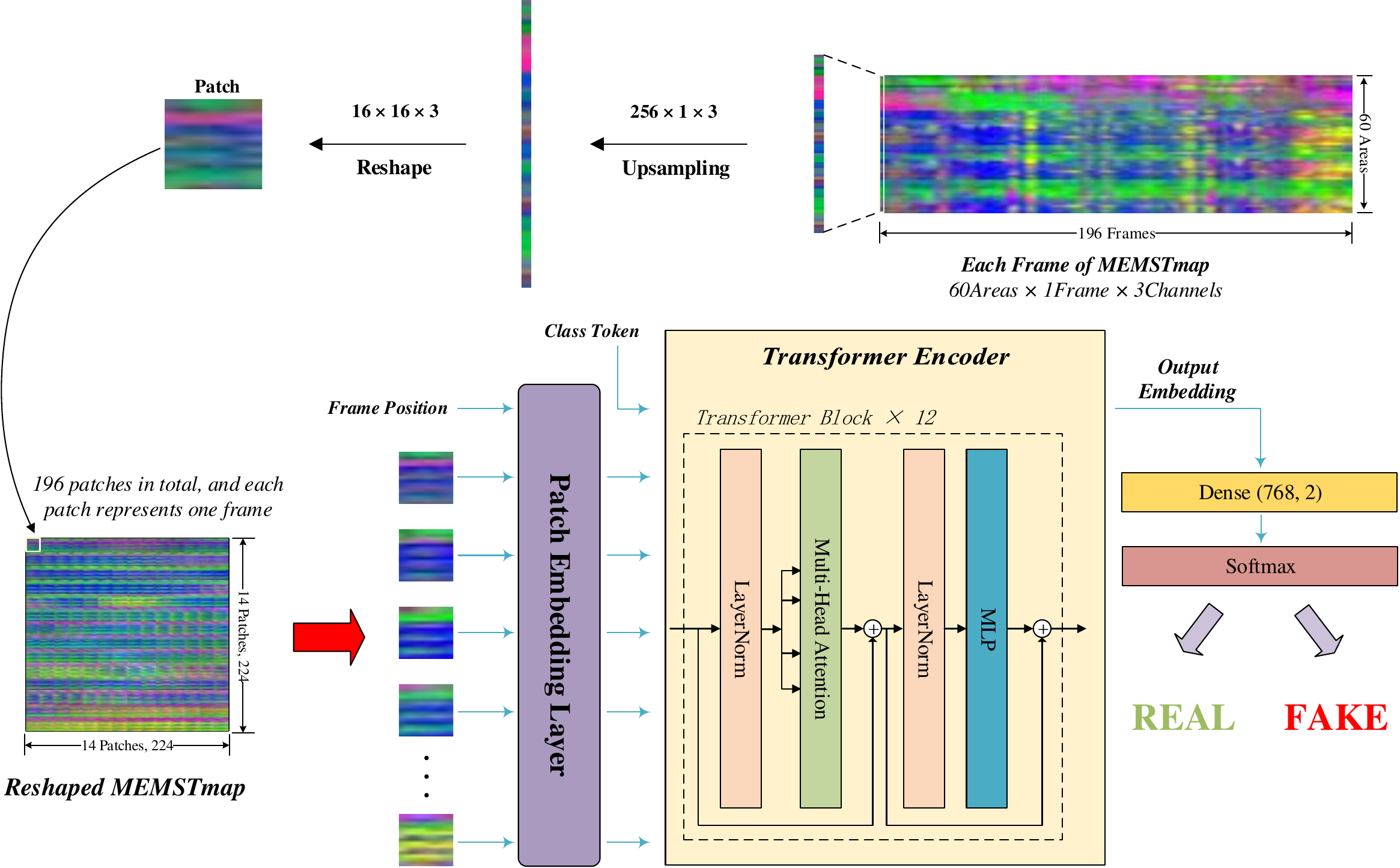}
	\end{center}
	\caption{\textbf{Overview of our proposed FakeTransformer.} In order to decompose the MEMSTmap into the input token of the vision Transformer, and ensure that the input embedding transformation will not destroy the temporal information hidden in each column of the spatial-temporal map when segment the picture into patches, we interpolate and reshape each column of MEMSTmap, encapsulate the spatial color information of each facial video frame into a patch, and input these patches and their corresponding frame positions into the Transformer encoder to learn the temporal and spatial pattern differences of facial pixel variations in real and forgery videos.}
	\label{fig:architecture}
\end{figure*}

\subsection{DeepFakes Detection Methods}
At present, the mainstream DeepFakes detection methods mainly have two directions. The first one is to detect the intra-frame image artifacts introduced during face synthesis or affine transformation, such as inconsistent head pose \cite{yang2019exposing}, color cues different from real cameras \cite{mccloskey2018detecting}, broken local PRNU patterns \cite{koopman2018detection,lukas2006digital} or detectable differences based on image quality measures (IQM) \cite{galbally2014face,korshunov2018deepfakes}. The other one is to detect inter-frame artifacts in an attempt to discover the artificial temporal inconsistency caused by frame by frame operation. As a representative, Sabir \etal \cite{sabir2019recurrent} proposed an RCN model to detect the temporal artifacts with inconsistent features across frames, and similarly, Guera and Delp \cite{guera2018deepfake} chose to use CNN and LSTM to create a time series descriptor to detect the authenticity of the video.

Ciftci \etal \cite{ciftci2020fakecatcher} first proposed to use physiological signals to detect DeepFakes, their statistics found that the time-frequency characteristics of rPPG signals extracted from synthetic videos are significantly different from those in real videos and they made a series of hand-craft descriptors to identify DeepFakes. In their follow-up work \cite{ciftci2020hearts}, they further used CNN to extract the differencs in rPPG signal and its spectrum, which improved the accuracy of detection.

\subsection{Remote Photoplethysmography}
Remote Photoplethysmography (rPPG) is a vision-based contactless remote physiological monitoring method that can seperate the physiological information of the human body (e.g. heart rate, respiration rate, BVP signal and $SpO_2$) from the facial pixel variations recorded by RGB cameras. Its technical focus is to separate physiologically related components from noise such as illumination, skin tone, and motion. Early traditional methods include blind source separation (BSS) methods based on PCA \cite{lewandowska2011measuring} and ICA \cite{poh2010advancements}, or CHROM \cite{de2013robust}, PBV \cite{de2014improved} and POS \cite{wang2016algorithmic} methods based on skin reflection models. However, most of these methods have strict requirements on video conditions and are deeply affected by compression and motion. Later works proposed some methods based on deep neural networks \cite{niu2020video,chen2018deepphys,yu2019remote}, which improved the situation to some extent.


\section{Proposed Method}
In this paper, with the perspective of modeling spatial-temporal representation, we hope to find the temporal and spatial difference between real videos and synthetic videos from the color sequence extracted from the face. Therefore, we propose FakeTransformer and the overview is shown in the figure \ref{fig:architecture}. Our workflow will be described in three parts, firstly, we give the analysis of the target task, then introduce the generation of MEMSTmap, and finally give the details of FakeTransformer.

\subsection{Analysis of Our Work}
Previous work has proved that the current generative model can not perfectly restore the color variation caused by the change of capillary blood volume. Therefore, part of the physiological information in the GAN-generated contents is destroyed. Combining the previous research results, we give the following hypotheses. 1) Due to the frame by frame face swapping operation in the DeepFakes generation pipeline, the physiological variations extracted from the face are not continuous in time sequence. 2) Human facial capillaries have a certain distribution pattern, which will cause the phase difference of periodic color signals in different facial areas in the heart beat cycle, and the GAN-generated face does not have this fixed pattern, so the physiological information in real videos and synthetic videos also has spatial differences.

The past physiological information-based DeepFakes detection methods often extract rPPG signals through traditional methods, and then find detectable differences from the time or frequency domain, however, these traditional methods often obtain physiological components from facial color sequence through linear dimension reduction methods such as PCA. Unfortunately, the error caused by manipulation is often not within the principal component, so it is easy to be eliminated. At the same time, rPPG signals based on manual extraction are easy to be affected by video factors such as illumination, skin color or motion, so that it is difficult to adapt to the actual Internet video conditions. In order to solve these problems, we extract the spatial average of each color channel from several face regions frame by frame to form a signal group, and hope to find the pattern difference between real video and forged video from the temporal and spatial variations of facial pixels.

Intuitively speaking, the spatial-temporal signal group can be regarded as an interdependent multivariate time-series, in which the spatial dimension (i.e. the variable dimension of the series) contains the physiological information of different facial positions, and the time dimension contains the characteristics of temporal variations between frames. Therefore, we can turn the problem into how to find the spatial and temporal patterns from the multivariate time-series. As we all know, in language or time-series tasks, the Transformer structure model can effectively capture the correlation between each token in a long sequence for its powerful global self-attention mechanism, so we choose to use Transformer to encode the given multivariate time-series and obtain a robust descriptor to expose spatial and temporal artifacts. Inspired by the practical experience of some advanced language models (e.g. Bert \cite{devlin2018bert}, GPT \cite{radford2018improving}), which training on large corpora and fine tuning on specific dataset, we choose to use the pre-trained Transformer encoder with excellent initialization parameters.

Each column of the spatial-temporal signal group (i.e. each frame in the video) can be regarded as an input word embedding of the Transformer after a certain linear transformation. However, due to the limited length of the pre-trained word vector and the output of the pixel value vector after the linear layer is uncontrollable, the word embedding is prone to overflow, so we choose to use the vision Transformer (ViT) \cite{dosovitskiy2020image} as our backbone because it has more suitable embedding transformation for pixel input. We convert the spatial-temporal signal groups into spatial-temporal maps and reshape them in units of columns to ensure that the time information will not be damaged during patch segmentation. It is worth mentioning that although we use the vision Transformer to learn the spatial-temporal map, the reshaped map does not have any image meaning. We encapsulate the spatial color information of each frame into a patch, and use the input embedding transformation of the vision Transformer to convert it into a form that can be accepted by the encoder, so essentially, our method is still modeling a multivariate sequence, aiming to learn the inter-frame temporal consistency and the intra-frame spatial coherence.

\subsection{Multi-Scale Eulerian Magnified Spatial-Temporal map}
The optical absorption variations of facial area caused by heartbeat is always subtle, and this fine signal is easy to be disturbed by lighting or motion. At the same time, only using the source video frame as the input of the detection model will bring unnecessary redundant information to the training, so as to reduce the accuracy. In order to solve this problem, we first decompose the original video into multiple Gaussian scales, filter the heart-rate related contents from the corresponding octave and enlarge it. Then the color signals of the facial region of the original video and physiological enhanced videos are extracted frame by frame. Finally, after normalization and segmentation, a fixed length spatial-temporal representation is formed as the input of our model.

Specifically, the generation pipeline of MEMSTmap is shown in Figure \ref{fig:MEMSTmap}. Firstly, we decompose the input original video sequence into three Gaussian spatial octaves, these contents are then temporal band-pass filtered in the range of heart rate frequency (0.75Hz to 3Hz). For each octave, we multiply it by an appropriate amplification factor to ensure that there is no overflow of pixel values when superimposing back to the original video. Then, considering that a large area of occlusion (hair, hat or headdress) often appears on the forehead, which affects the extraction of skin information, we divide the remaining facial area into 15 RoIs through landmarks. For each RoI of each video sequence, we calculate the spatial average of Y, U and V color channels (As described in work \cite{niu2019rhythmnet}, YUV color space is effective in representing the physiological information than RGB color space) frame by frame to form three temporal signals. Finally, we normalize the signals, take 196 frames as the window length and 0.5s as the sliding length to form several 3-channels (YUV) MEMSTmaps with a width of 196 and a height of 60 (4 videos × 15 RoIs).

\subsection{FakeTransformer}
In this section, we will detail the pipeline of our proposed FakeTransformer. As we mentioned in 3.1, we intend to use MEMSTmap to fine-tune the parameters of the pre-trained vision Transformer. Each column of MEMSTmap represents the color information extract from each RoI in the original video and the enhanced video. If we send it directly to the vision Transformer for further learning, the forced segmentation will destroy the time information hidden in each column. Therefore, we reshape each column of MEMSTmap and encapsulate the spatial color information of each frame into an independent patch to meet the input requirements of the vision Transformer, so that retain important inter-frame temporal features without modifying the input embedding transformation of the pre-trained model.

Specifically, we use a pre-trained vision Transformer with an input image size of 224 and a patch size of 16. So we interpolate each column of MEMSTmap to 256, and then reshape it to 16 by 16 to form a input patch. Then, we orderly input each patch and its corresponding frame position into the Transformer encoder, and obtain a 762 dimensional spatio-time descriptor from the output position corresponding to the class token. Finally, we use a dense layer to purify the feature embedding and use a softmax layer to output the probability of judging whether the MEMSTmap is real or fake.

\section{Experiment and Analysis}
In this section, we will give the implementation details of our proposed FakeTransformer, the results of comparison with the baseline method on each sub-datasets, and a series of ablation study result.

\subsection{Training Details}
\subsubsection{DeepFakes Datasets and Pre-processing}
Our work mainly uses FaceForensics++ dataset \cite{rossler2019faceforensics++} to train and test our model, and randomly selects a certain number of real videos and forged videos from DeepFake Detection dataset to form a test set to verify the cross-domain generalization ability of our approach. It should be noted that the DeepFake Detection dataset was first released by Google and now has been incorporated into FaceForensics++.

The FaceForensics++ dataset consists of real videos and synthetic videos from four different generation sources (i.e. DeepFakes, Face2Face, FaceSwap and NeuralTextures). The number of videos of each type is 1000. We establish sub datasets for each kind of forged videos, and constructed training sets, validation sets and test sets in the proportion of 8:1:1. In addition, due to the video length in FaceSwap and NeuralTextures is slightly shorter, in order to ensure the balance of samples, we take the first $70\%$ of the real videos to generate MEMSTmap when preparing the corresponding sub-datasets. The DeepFake Detection dataset consists of 363 real videos and 3068 synthetic videos. We randomly selected 100 real videos and 100 synthetic videos to form our cross data domain test set.

For each video, we use Dlib \cite{kazemi2014one} to detect the face position and locate the landmarks. If a video has more than 10 frames without finding a face, the video will be discarded. We convert all videos to the form of MEMSTmap, in which the training set and validation set are in the unit of map, and the test set is in the unit of video. The number of MEMSTmap for each sub-dataset is shown in Table \ref{tab:table1}.

\subsubsection{Training Settings}
In our training process, we choose the cross entropy function as the classification loss function and Adam \cite{kingma2014adam} as the optimizer. We use the base version pre-trained ViT with the input image size of 224 and the patch size of 16 as our backbone, and fine tune the parameters at the learning rate of 0.00005 for total 60 epochs. In order to prevent over fitting, we add dropout \cite{srivastava2014dropout} on the output embedding of ViT with a probability of 0.1.

\begin{table}[t]
	\Large
	\centering
	\renewcommand\arraystretch{1.6} 
	\begin{center}
		\resizebox{0.48\textwidth}{!}{%
			\begin{tabular}{|c|c|cc|cc|cc|}
				\hline
				\rowcolor[HTML]{EFEFEF} 
				\cellcolor[HTML]{EFEFEF} &
				\cellcolor[HTML]{EFEFEF} &
				\multicolumn{2}{c|}{\cellcolor[HTML]{EFEFEF}\textbf{Train}} &
				\multicolumn{2}{c|}{\cellcolor[HTML]{EFEFEF}\textbf{Validation}} &
				\multicolumn{2}{c|}{\cellcolor[HTML]{EFEFEF}\textbf{Test}} \\ \cline{3-8} 
				\rowcolor[HTML]{EFEFEF} 
				\multirow{-2}{*}{\cellcolor[HTML]{EFEFEF}\textbf{Sub-Dataset}} &
				\multirow{-2}{*}{\cellcolor[HTML]{EFEFEF}\textbf{Total}} &
				\textbf{Real} &
				\textbf{Fake} &
				\textbf{Real} &
				\textbf{Fake} &
				\textbf{Real} &
				\textbf{Fake} \\ \hline
				\textbf{DeepFakes}          & 44430 & 18370 & 18085 & 2011 & 1935 & 2058 & 1971 \\
				\textbf{Face2Face}          & 44851 & 18370 & 18323 & 2011 & 1982 & 2058 & 2107 \\
				\textbf{FaceSwap}           & 30669 & 12943 & 12099 & 1428 & 1365 & 1459 & 1375 \\
				\textbf{NeuralTextures}     & 30718 & 12943 & 12141 & 1428 & 1365 & 1459 & 1383 \\ \hline
				\textbf{DeepFake Detection} & 9547  & -     & -     & -    & -    & 5091 & 4456 \\ \hline
			\end{tabular}%
		}
	\end{center}
	\caption{We list the number of MEMSTmap of Real videos and Fake videos in each sub-dataset, DeepFake Detection is only used as the test set, so there are no training and validation samples.}
	\label{tab:table1}
\end{table}

\begin{table*}[t]
	
	\centering
	\renewcommand\arraystretch{1.4} 
	\resizebox{0.9\textwidth}{!}{%
		\begin{tabular}{|c|cccc|c|}
			\hline
			\rowcolor[HTML]{EFEFEF} 
			\cellcolor[HTML]{EFEFEF} & \multicolumn{5}{c|}{\cellcolor[HTML]{EFEFEF}\textbf{Sub-Dataset}}            \\ \cline{2-6} 
			\rowcolor[HTML]{EFEFEF} 
			\multirow{-2}{*}{\cellcolor[HTML]{EFEFEF}\textbf{Baseline}} &
			\textbf{DeepFakes} &
			\textbf{Face2Face} &
			\textbf{FaceSwap} &
			\textbf{NeuralTextures} &
			\textbf{DeepFake Detection} \\ \hline
			Inception V3 \cite{szegedy2016rethinking}             & 83.76\%          & 84.50\%          & 82.91\%          & 69.00\%          & 56.50\% \\
			MesoNet \cite{afchar2018mesonet}                 & 95.94\%          & 96.50\%          & 96.98\%          & 79.50\%          & 66.50\% \\
			Xception \cite{chollet2017xception}                 & \textbf{98.98\%} & 97.50\%          & 96.48\%          & 83.00\%          & 62.50\% \\
			PPG cell \cite{ciftci2020hearts}                & 94.42\%          & 93.00\%          & 93.47\%          & 77.00\%          & 64.00\% \\ \hline
			\textbf{ours}            & \textbf{98.98\%}          & \textbf{98.50\%} & \textbf{98.49\%} & \textbf{86.50\%} &  \textbf{68.00\%}\\ \hline
		\end{tabular}%
	}
	\caption{The binary classification accuracy of our method compared with other baseline methods on each sub-dataset, we highlight the optimal results in the table. We also list the cross-domain detection results of each model on the DeepFake Detection dataset.}
	\label{tab:table2}
\end{table*}

\subsection{Baseline Comparision on Accuracy}
\subsubsection{Selection of Baseline}
We selected Xception \cite{chollet2017xception}, Inception V3 \cite{szegedy2016rethinking}, MesoNet \cite{afchar2018mesonet} and PPG cell \cite{ciftci2020hearts} as our baseline for comparison. These methods have well performance in the DeepFakes detection task. For Xception and Inception V3, we modified the final dense layer to adapt to our classification target. For MesoNet, we use the source code provided by the author and retrain the model. For PPG cell method, we re-implemented the model with the length of 64 frames according to the paper. 

It should be noted that since Xception, Inception V3 and MesoNet use image as input, so we need to make some adjustments to the training and testing process in our video task. Specifically, we split the video into frames and extract the facial region. Similarly, these frames are divided into training, validation and test sets according to the video unit. When testing the model, we use the network to predict each frame in the video, count the number of frames judged as real or fake, and finally determine the attribute of the video by majority voting.

\subsubsection{Accuracy on FaceForensics++}
When verifying our model, since a video contains more than one MEMSTmap, we  make prediction for the authenticity of each map, and then majority vote for the result of the video. If the number of real predictions and fake predictions are the same, the predicted output of each MEMSTmap is counted and averaged, and the video attribute with high probability is taken as the final result.

We list the binary detection accuracy of our method compared with other baselines on each sub-dataset of FaceForensics++ in the Table \ref{tab:table2}, It can be seen that, in general, our method has better performance than other baseline methods. Specifically, our FakeTransformer achieves $98.50\%$, $98.49\%$ and $86.50\%$ accuracy on Face2Face, FaceSwap and NeuralTextures respectively, which are higher than other baselines. It is worth mentioning that our method is far better than the second in the more realistic NeuralTextures sub-dataset (i.e. $83.00\%$ of Xception). On the DeepFakes sub-dataset, both our approach and Xception achieve $98.98\%$ accuracy and has better performance than other models. Compared with PPG cell, which is also a physiological information-based approach, our model shows higher detection accuracy on all sub-datasets, confirming that our model has SOTA performance in similar methods.

\subsection{Robustness Experiment}
\subsubsection{Cross-Dataset Robustness}
In order to verify the cross-data domain robustness of our model, we randomly select 200 videos (100 real videos and 100 synthetic videos respectively) from the DeepFake Detection dataset and form a independent test set. For the model to be verified, we uniformly train with the DeepFakes sub-dataset, and use all videos to train the models, without separating the validation set and test set.

The results for cross-data experiment are also list in Table \ref{tab:table2}. Obviously, the cross-data domain robustness of inception V3 is the worst among all baselines, and the accuracy on the test set of DeepFake Detection is only $56.5\%$. Although Xception shows strong performance in binary classification tasks, due to its complex network structure, it may show a certain over-fitting for cross-data domain samples, with an accuracy of only $62.5\%$. On the contrary, the lightweight network MesoNet shows better cross domain robustness, and has reached an accuracy of $66.5\%$. Our method is good at exposing the widespread physiological artifacts in forged videos, and achieves the highest cross-data domain classification accuracy compared with other baseline, that is, $68\%$, which is $4\%$ higher than the similar PPG cell model based on rPPG signal and its spectrum.

\begin{figure}[t]
	\begin{center}
		\includegraphics[width=0.95\linewidth]{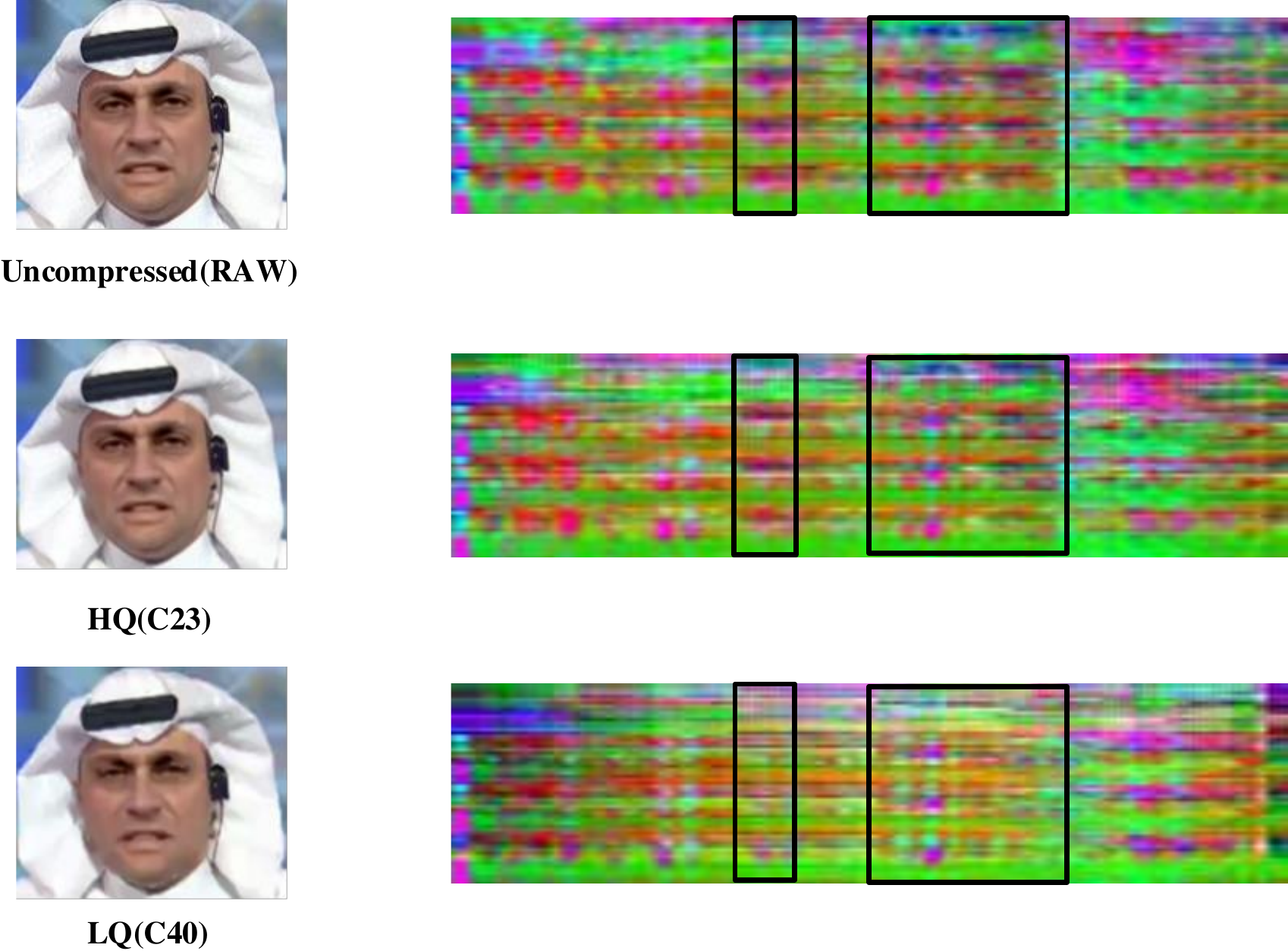}
	\end{center}
	\caption{We show the differences of MEMSTmap under different compression levels and highlight several obvious inconsistent areas. It can be seen that the variations of the compressed MEMSTmap are reduced both in temporal and spatial dimensions.}
	\label{fig:compress}
\end{figure}

\subsubsection{Video Compression Robustness}
physiological information-based Deepfake detection models are often seriously affected by video compression, which is due to the limitations of remote physiological monitoring method. People often compress the video with large storage space to ensure that the video can be transmitted through the channel with limited bandwidth. In this process, a lot of facial details will be deleted and replaced, and adjacent pixel blocks will be combined as much as possible to reduce the amount of information carried by the video. Therefore, the color variation signal extracted from the face will be destroyed, which will affect the judgment of the model on the spatial-temporal sequence.

We show the differences of MEMSTmap under different compression levels in figure \ref{fig:compress}. Due to the facial details are downsampled, the color differences between different regions are eliminated, and the colors of each region of the face tend to be consistent, which also destroys the variations in temporal dimension.

We use three different compression levels (i.e. raw, C23 and C40) DeepFakes sub-datasets to implement video compression robustness experiments, and compare with PPG cell, which is also based on physiological information. The results are shown in the Table \ref{tab:table3}. It can be seen that our method has less accuracy degradation than PPG cell when facing highly compressed video, Specifically, for C23 video, our method accuracy decreases by $3.07\%$, PPG cell decreases by $5.39\%$, and for C40 video, our method accuracy decreases by $11.28\%$, PPG cell decreases by $16.67\%$. 

We believe that the past physiological information-based detection method extracted the rPPG signal manually. When performing PCA on the pixel sequence, the introduced compression artifact will be amplified, so that the detectable cues that can distinguish between real videos and synthetic videos may be ignored. Our method is not limited to extracting physiological signals, instead, we uses Transformer to model the temporal-spatial sequence of facial pixel variations, which reduces the impact of compression on the detection accuracy.

\begin{table}[]
	\centering
	\renewcommand\arraystretch{1.4} 
	\tiny
	\resizebox{0.45\textwidth}{!}{%
		\begin{tabular}{|c|ccc|}
			\hline
			\rowcolor[HTML]{EFEFEF} 
			\cellcolor[HTML]{EFEFEF}                                  & \multicolumn{3}{c|}{\cellcolor[HTML]{EFEFEF}\textbf{Compression Level}} \\ \cline{2-4} 
			\rowcolor[HTML]{EFEFEF} 
			\multirow{-2}{*}{\cellcolor[HTML]{EFEFEF}\textbf{Method}} & \textbf{raw}           & \textbf{C23}           & \textbf{C40}          \\ \hline
			PPG cell      & 94.42\% & 89.33\% & 78.68\% \\ \hline
			\textbf{ours} & 98.98\% & 95.94\% & 87.82\% \\ \hline
		\end{tabular}%
	}
	\caption{Experiment results for video compression robustness on DeepFakes sub-dataset.}
	\label{tab:table3}
\end{table}

\subsection{Ablation Study}
In this section, we will do ablation study on our method from the following three aspects: 1) effectiveness of Vision Transformer (ViT), 2) effectiveness of  Eulerian Video Magnification and 3) effectiveness of reshaping the map. Among them, in order to verify the effectiveness of ViT, we use ResNet18 (RN) to replace Transformer as the backbone to learn from MEMSTmap. In order to verify the effectiveness of Eulerian Video Magnification, we use the MSTmap proposed in work \cite{niu2020video} instead of MEMSTmap as learning samples. The spatial-temporal map proposed by Niu \etal does not contain enlarged physiological information. In order to verify the effectiveness of reshaping the map, we use ViT to train and test with the unmodified MEMSTmap. The results of ablation experiment are shown in Table \ref{tab:table4}.

By comparing the first and third rows of the table, the detection accuracy based on ResNet18 is much lower than that based on ViT, which proves that Transformer's global self attention mechanism plays an important role in finding spatial-temporal artifacts caused by forgery. By comparing the second and fourth rows of the table, it can be seen that the detection accuracy of MEMSTmap with enhanced facial physiological variation information is better than that of MSTmap without enhancement, which proves the effectiveness of enhancing physiological expression on multiple octave scales. By comparing the third and fourth rows of the table, in general, the reshaped MEMSTmap has higher detection accuracy than the unmodified map, which proves that the retained time information is helpful for the neural network to find temporal inconsistency in the manipulation videos.

\begin{table}[t]
	\centering
	\LARGE
	\renewcommand\arraystretch{1.4}  
	\resizebox{0.5\textwidth}{!}{%
		\begin{tabular}{|c|cccc|}
			\hline
			\rowcolor[HTML]{EFEFEF} 
			\cellcolor[HTML]{EFEFEF}                                    & \multicolumn{4}{c|}{\cellcolor[HTML]{EFEFEF}\textbf{Sub-Dataset}}                     \\ \cline{2-5} 
			\rowcolor[HTML]{EFEFEF} 
			\multirow{-2}{*}{\cellcolor[HTML]{EFEFEF}\textbf{Settings}} & \textbf{DeepFakes} & \textbf{Face2Face} & \textbf{FaceSwap} & \textbf{NeuralTextures} \\ \hline
			RN + MEMSTmap     & 84.26\% & 83.00\% & 85.43\% & 74.50\% \\
			ViT + MSTmap(r)   & 92.38\% & 91.00\% & 92.46\% & 79.00\% \\
			ViT + MEMSTmap    & 96.95\% & 97.50\% & 96.98\% & 87.50\% \\
			ViT + MEMSTmap(r) & 98.98\% & 98.50\% & 98.49\% & 86.50\% \\ \hline
		\end{tabular}%
	}
	\caption{The ablation study of the model proposed in this paper by progressively setting ResNet18 (RN), Vision Transformer (ViT), MSTmap and reshape (r) operation.}
	\label{tab:table4}
\end{table}

\section{Conclusion}
In this paper, we propose a face manipulation detection method, which we named FakeTransformer. For the first time, we regard the facial pixel variation sequence as a multivariate time-series, and look for detectable artifacts that can distinguish face forgery from the physiologically enhanced spatial-temporal representation through the Transformer's powerful self-attention mechanism. We have verified the effectiveness and cross-data domain robustness of our model on FaceForensics++ and DeepFake Detection datasets, and both of them have achieved excellent performance.

Due to the inherent limitations of facial physiological feature extraction methods, our model will reduce the expansion ability for highly compressed video. Therefore, In our follow-up work, we intend to combine the measurement of facial region trajectory features to improve the ability of resisting video compression.

{\small
\bibliographystyle{ieee}
\bibliography{egbib}

\begin{thebibliography}{10}\itemsep=-1pt

\bibitem{afchar2018mesonet}
D.~Afchar, V.~Nozick, J.~Yamagishi, and I.~Echizen.
\newblock Mesonet: a compact facial video forgery detection network.
\newblock In {\em 2018 IEEE International Workshop on Information Forensics and
  Security (WIFS)}, pages 1--7. IEEE, 2018.

\bibitem{amarasinghe2018toward}
K.~Amarasinghe, K.~Kenney, and M.~Manic.
\newblock Toward explainable deep neural network based anomaly detection.
\newblock In {\em 2018 11th International Conference on Human System
  Interaction (HSI)}, pages 311--317. IEEE, 2018.

\bibitem{bagnall2015time}
A.~Bagnall, J.~Lines, J.~Hills, and A.~Bostrom.
\newblock Time-series classification with cote: the collective of
  transformation-based ensembles.
\newblock {\em IEEE Transactions on Knowledge and Data Engineering},
  27(9):2522--2535, 2015.

\bibitem{basumallik2019packet}
S.~Basumallik, R.~Ma, and S.~Eftekharnejad.
\newblock Packet-data anomaly detection in pmu-based state estimator using
  convolutional neural network.
\newblock {\em International Journal of Electrical Power \& Energy Systems},
  107:690--702, 2019.

\bibitem{chen2018deepphys}
W.~Chen and D.~McDuff.
\newblock Deepphys: Video-based physiological measurement using convolutional
  attention networks.
\newblock In {\em Proceedings of the European Conference on Computer Vision
  (ECCV)}, pages 349--365, 2018.

\bibitem{chollet2017xception}
F.~Chollet.
\newblock Xception: Deep learning with depthwise separable convolutions.
\newblock In {\em Proceedings of the IEEE conference on computer vision and
  pattern recognition}, pages 1251--1258, 2017.

\bibitem{ciftci2020fakecatcher}
U.~A. Ciftci, I.~Demir, and L.~Yin.
\newblock Fakecatcher: Detection of synthetic portrait videos using biological
  signals.
\newblock {\em IEEE Transactions on Pattern Analysis and Machine Intelligence},
  2020.

\bibitem{ciftci2020hearts}
U.~A. Ciftci, I.~Demir, and L.~Yin.
\newblock How do the hearts of deep fakes beat? deep fake source detection via
  interpreting residuals with biological signals.
\newblock In {\em 2020 IEEE International Joint Conference on Biometrics
  (IJCB)}, pages 1--10. IEEE, 2020.

\bibitem{de2013robust}
G.~De~Haan and V.~Jeanne.
\newblock Robust pulse rate from chrominance-based rppg.
\newblock {\em IEEE Transactions on Biomedical Engineering}, 60(10):2878--2886,
  2013.

\bibitem{de2014improved}
G.~De~Haan and A.~Van~Leest.
\newblock Improved motion robustness of remote-ppg by using the blood volume
  pulse signature.
\newblock {\em Physiological measurement}, 35(9):1913, 2014.

\bibitem{devlin2018bert}
J.~Devlin, M.-W. Chang, K.~Lee, and K.~Toutanova.
\newblock Bert: Pre-training of deep bidirectional transformers for language
  understanding.
\newblock {\em arXiv preprint arXiv:1810.04805}, 2018.

\bibitem{dosovitskiy2020image}
A.~Dosovitskiy, L.~Beyer, A.~Kolesnikov, D.~Weissenborn, X.~Zhai,
  T.~Unterthiner, M.~Dehghani, M.~Minderer, G.~Heigold, S.~Gelly, et~al.
\newblock An image is worth 16x16 words: Transformers for image recognition at
  scale.
\newblock {\em arXiv preprint arXiv:2010.11929}, 2020.

\bibitem{galbally2014face}
J.~Galbally and S.~Marcel.
\newblock Face anti-spoofing based on general image quality assessment.
\newblock In {\em 2014 22nd international conference on pattern recognition},
  pages 1173--1178. IEEE, 2014.

\bibitem{geng2018cost}
Y.~Geng and X.~Luo.
\newblock Cost-sensitive convolution based neural networks for imbalanced
  time-series classification.
\newblock {\em arXiv preprint arXiv:1801.04396}, 2018.

\bibitem{guera2018deepfake}
D.~G{\"u}era and E.~J. Delp.
\newblock Deepfake video detection using recurrent neural networks.
\newblock In {\em 2018 15th IEEE international conference on advanced video and
  signal based surveillance (AVSS)}, pages 1--6. IEEE, 2018.

\bibitem{kate2016using}
R.~J. Kate.
\newblock Using dynamic time warping distances as features for improved time
  series classification.
\newblock {\em Data Mining and Knowledge Discovery}, 30(2):283--312, 2016.

\bibitem{kazemi2014one}
V.~Kazemi and J.~Sullivan.
\newblock One millisecond face alignment with an ensemble of regression trees.
\newblock In {\em Proceedings of the IEEE conference on computer vision and
  pattern recognition}, pages 1867--1874, 2014.

\bibitem{kingma2014adam}
D.~P. Kingma and J.~Ba.
\newblock Adam: A method for stochastic optimization.
\newblock {\em arXiv preprint arXiv:1412.6980}, 2014.

\bibitem{koopman2018detection}
M.~Koopman, A.~M. Rodriguez, and Z.~Geradts.
\newblock Detection of deepfake video manipulation.
\newblock In {\em The 20th Irish machine vision and image processing conference
  (IMVIP)}, pages 133--136, 2018.

\bibitem{korshunov2018deepfakes}
P.~Korshunov and S.~Marcel.
\newblock Deepfakes: a new threat to face recognition? assessment and
  detection.
\newblock {\em arXiv preprint arXiv:1812.08685}, 2018.

\bibitem{lewandowska2011measuring}
M.~Lewandowska, J.~Rumi{\'n}ski, T.~Kocejko, and J.~Nowak.
\newblock Measuring pulse rate with a webcam—a non-contact method for
  evaluating cardiac activity.
\newblock In {\em 2011 federated conference on computer science and information
  systems (FedCSIS)}, pages 405--410. IEEE, 2011.

\bibitem{lines2016hive}
J.~Lines, S.~Taylor, and A.~Bagnall.
\newblock Hive-cote: The hierarchical vote collective of transformation-based
  ensembles for time series classification.
\newblock In {\em 2016 IEEE 16th international conference on data mining
  (ICDM)}, pages 1041--1046. IEEE, 2016.

\bibitem{liu2005motion}
C.~Liu, A.~Torralba, W.~T. Freeman, F.~Durand, and E.~H. Adelson.
\newblock Motion magnification.
\newblock {\em ACM transactions on graphics (TOG)}, 24(3):519--526, 2005.

\bibitem{lukas2006digital}
J.~Lukas, J.~Fridrich, and M.~Goljan.
\newblock Digital camera identification from sensor pattern noise.
\newblock {\em IEEE Transactions on Information Forensics and Security},
  1(2):205--214, 2006.

\bibitem{lyu2020deepfake}
S.~Lyu.
\newblock Deepfake detection: Current challenges and next steps.
\newblock In {\em 2020 IEEE International Conference on Multimedia \& Expo
  Workshops (ICMEW)}, pages 1--6. IEEE, 2020.

\bibitem{malhotra2017timenet}
P.~Malhotra, V.~TV, L.~Vig, P.~Agarwal, and G.~Shroff.
\newblock Timenet: Pre-trained deep recurrent neural network for time series
  classification.
\newblock {\em arXiv preprint arXiv:1706.08838}, 2017.

\bibitem{mccloskey2018detecting}
S.~McCloskey and M.~Albright.
\newblock Detecting gan-generated imagery using color cues.
\newblock {\em arXiv preprint arXiv:1812.08247}, 2018.

\bibitem{mirsky2021creation}
Y.~Mirsky and W.~Lee.
\newblock The creation and detection of deepfakes: A survey.
\newblock {\em ACM Computing Surveys (CSUR)}, 54(1):1--41, 2021.

\bibitem{nguyen2019deep}
T.~T. Nguyen, C.~M. Nguyen, D.~T. Nguyen, D.~T. Nguyen, and S.~Nahavandi.
\newblock Deep learning for deepfakes creation and detection: A survey.
\newblock {\em arXiv preprint arXiv:1909.11573}, 2019.

\bibitem{niu2019rhythmnet}
X.~Niu, S.~Shan, H.~Han, and X.~Chen.
\newblock Rhythmnet: End-to-end heart rate estimation from face via
  spatial-temporal representation.
\newblock {\em IEEE Transactions on Image Processing}, 29:2409--2423, 2019.

\bibitem{niu2020video}
X.~Niu, Z.~Yu, H.~Han, X.~Li, S.~Shan, and G.~Zhao.
\newblock Video-based remote physiological measurement via cross-verified
  feature disentangling.
\newblock In {\em European Conference on Computer Vision}, pages 295--310.
  Springer, 2020.

\bibitem{oh2018learning}
T.-H. Oh, R.~Jaroensri, C.~Kim, M.~Elgharib, F.~Durand, W.~T. Freeman, and
  W.~Matusik.
\newblock Learning-based video motion magnification.
\newblock In {\em Proceedings of the European Conference on Computer Vision
  (ECCV)}, pages 633--648, 2018.

\bibitem{poh2010advancements}
M.-Z. Poh, D.~J. McDuff, and R.~W. Picard.
\newblock Advancements in noncontact, multiparameter physiological measurements
  using a webcam.
\newblock {\em IEEE transactions on biomedical engineering}, 58(1):7--11, 2010.

\bibitem{radford2018improving}
A.~Radford, K.~Narasimhan, T.~Salimans, and I.~Sutskever.
\newblock Improving language understanding by generative pre-training.
\newblock 2018.

\bibitem{rossler2019faceforensics++}
A.~Rossler, D.~Cozzolino, L.~Verdoliva, C.~Riess, J.~Thies, and M.~Nie{\ss}ner.
\newblock Faceforensics++: Learning to detect manipulated facial images.
\newblock In {\em Proceedings of the IEEE/CVF International Conference on
  Computer Vision}, pages 1--11, 2019.

\bibitem{sabir2019recurrent}
E.~Sabir, J.~Cheng, A.~Jaiswal, W.~AbdAlmageed, I.~Masi, and P.~Natarajan.
\newblock Recurrent convolutional strategies for face manipulation detection in
  videos.
\newblock {\em Interfaces (GUI)}, 3(1):80--87, 2019.

\bibitem{srivastava2014dropout}
N.~Srivastava, G.~Hinton, A.~Krizhevsky, I.~Sutskever, and R.~Salakhutdinov.
\newblock Dropout: a simple way to prevent neural networks from overfitting.
\newblock {\em The journal of machine learning research}, 15(1):1929--1958,
  2014.

\bibitem{strodthoff2019detecting}
N.~Strodthoff and C.~Strodthoff.
\newblock Detecting and interpreting myocardial infarction using fully
  convolutional neural networks.
\newblock {\em Physiological measurement}, 40(1):015001, 2019.

\bibitem{szegedy2016rethinking}
C.~Szegedy, V.~Vanhoucke, S.~Ioffe, J.~Shlens, and Z.~Wojna.
\newblock Rethinking the inception architecture for computer vision.
\newblock In {\em Proceedings of the IEEE conference on computer vision and
  pattern recognition}, pages 2818--2826, 2016.

\bibitem{vaswani2017attention}
A.~Vaswani, N.~Shazeer, N.~Parmar, J.~Uszkoreit, L.~Jones, A.~N. Gomez,
  {\L}.~Kaiser, and I.~Polosukhin.
\newblock Attention is all you need.
\newblock In {\em Advances in neural information processing systems}, pages
  5998--6008, 2017.

\bibitem{verdoliva2020media}
L.~Verdoliva.
\newblock Media forensics and deepfakes: an overview.
\newblock {\em IEEE Journal of Selected Topics in Signal Processing},
  14(5):910--932, 2020.

\bibitem{wadhwa2013phase}
N.~Wadhwa, M.~Rubinstein, F.~Durand, and W.~T. Freeman.
\newblock Phase-based video motion processing.
\newblock {\em ACM Transactions on Graphics (TOG)}, 32(4):1--10, 2013.

\bibitem{wadhwa2014riesz}
N.~Wadhwa, M.~Rubinstein, F.~Durand, and W.~T. Freeman.
\newblock Riesz pyramids for fast phase-based video magnification.
\newblock In {\em 2014 IEEE International Conference on Computational
  Photography (ICCP)}, pages 1--10. IEEE, 2014.

\bibitem{wang2016algorithmic}
W.~Wang, A.~C. den Brinker, S.~Stuijk, and G.~De~Haan.
\newblock Algorithmic principles of remote ppg.
\newblock {\em IEEE Transactions on Biomedical Engineering}, 64(7):1479--1491,
  2016.

\bibitem{wu2012eulerian}
H.-Y. Wu, M.~Rubinstein, E.~Shih, J.~Guttag, F.~Durand, and W.~Freeman.
\newblock Eulerian video magnification for revealing subtle changes in the
  world.
\newblock {\em ACM transactions on graphics (TOG)}, 31(4):1--8, 2012.

\bibitem{yang2019exposing}
X.~Yang, Y.~Li, and S.~Lyu.
\newblock Exposing deep fakes using inconsistent head poses.
\newblock In {\em ICASSP 2019-2019 IEEE International Conference on Acoustics,
  Speech and Signal Processing (ICASSP)}, pages 8261--8265. IEEE, 2019.

\bibitem{yu2019remote}
Z.~Yu, W.~Peng, X.~Li, X.~Hong, and G.~Zhao.
\newblock Remote heart rate measurement from highly compressed facial videos:
  an end-to-end deep learning solution with video enhancement.
\newblock In {\em Proceedings of the IEEE/CVF International Conference on
  Computer Vision}, pages 151--160, 2019.

\bibitem{zhao2018novel}
C.~Zhao, C.-L. Lin, W.~Chen, and Z.~Li.
\newblock A novel framework for remote photoplethysmography pulse extraction on
  compressed videos.
\newblock In {\em Proceedings of the IEEE Conference on Computer Vision and
  Pattern Recognition Workshops}, pages 1299--1308, 2018.

\bibitem{zhou2021informer}
H.~Zhou, S.~Zhang, J.~Peng, S.~Zhang, J.~Li, H.~Xiong, and W.~Zhang.
\newblock Informer: Beyond efficient transformer for long sequence time-series
  forecasting.
\newblock In {\em Proceedings of AAAI}, 2021.

\end{thebibliography}
}

\end{document}